\documentclass{article}

\usepackage{microtype}
\usepackage{graphicx}
\usepackage{subfigure}
\usepackage{booktabs} 
\usepackage{multirow}
\usepackage{amsmath}
\usepackage{hyperref}

\usepackage{amsfonts}
\usepackage[accepted]{icml2020}

\usepackage{color}
\usepackage{array}

\newcolumntype{H}{>{\setbox0=\hbox\bgroup}c<{\egroup}@{}}

\icmltitlerunning{Few-shot link prediction via graph neural networks}
\usepackage{comment}
\newcommand{\specialcell}[2][c]{%
  \begin{tabular}[#1]{@{}c@{}}#2\end{tabular}}
\begin{document}

\twocolumn[
\icmltitle{Few-shot link prediction via graph neural networks for Covid-19 drug-repurposing}




\begin{icmlauthorlist}
\icmlauthor{Vassilis N. Ioannidis}{am}
\icmlauthor{Da Zheng}{am}
\icmlauthor{George Karypis}{am}
\end{icmlauthorlist}

\icmlaffiliation{am}{Amazon Web Services AI Lab, Palo Alto, California, USA}

\icmlcorrespondingauthor{Vassilis N. Ioannidis}{ioann006@umn.edu}

\icmlkeywords{Machine Learning, ICML}

\vskip 0.3in
]



\printAffiliationsAndNotice{}  

\begin{abstract}
Predicting interactions among heterogenous graph structured data has numerous applications such as knowledge graph completion, recommendation systems and drug discovery. Often times, the links to be predicted belong to rare types such as the case in repurposing drugs for novel diseases. This motivates the task of few-shot link prediction. Typically, GCNs are ill-equipped in learning such rare link types since the relation embedding is not learned in an inductive fashion. This paper proposes an inductive RGCN for learning informative relation embeddings even in the few-shot learning regime.   The proposed inductive model significantly outperforms the RGCN and state-of-the-art KGE models in few-shot learning tasks. Furthermore, we apply our method on the drug-repurposing knowledge graph (DRKG) for discovering drugs for Covid-19. We pose the drug discovery task as link prediction and learn embeddings for the biological entities that partake in the DRKG. Our initial results corroborate that several drugs used in clinical trials were identified as possible drug candidates.  The method in this paper are implemented using the efficient  deep graph learning (DGL)~\cite{wang2019deep}. 
\end{abstract}

\section{Introduction}
The timeline of the Covid-19 pandemic showcases the dire need for fast development of effective treatments for new diseases. Drug-repurposing is a drug discovery strategy from existing drugs that significantly shortens the time and reduces the cost compared to de~novo drug discovery~\cite{sertkaya2014examination,avorn20152,setoain2015nffinder}. Drug-repurposing leverages the fact that common molecular pathways contribute to different diseases and hence some drugs may be reused~\cite{ashburn2004drug}.

Drug-repurposing relies on identifying novel interactions among biological entities like genes and compounds and can be posed as a link prediction task over a biological network. Several machine learning approaches have been developed for addressing the drug-repurposingtask for Covid-19; see e.g.~\cite{gramatica2014graph, zhou2020network, udrescu2016clustering,drkg2020}. Towards assisting such machine learning techniques~\cite{drkg2020} created a comprehensive biological knowledge graph relating genes, compounds, diseases, biological processes, side effects and symptoms termed Drug Repurposing Knowledge Graph (DRKG).

However, for novel diseases like Covid-19 only a few interactions are available among viral proteins and possible chemical compounds that may inhibit the related genes. This motivates the framework of few-shot link prediction, where a certain edge type is rare and the model is called to make predictions on the particular edge type.

\subsection{Related works}
Link prediction has been addressed by several works in the context of knowledge-graph (KG) completion. These models rely on embedding the nodes and edges of the KG to a vector space and then train by maximizing the score for existing edges in the KGs; see e.g.,~\cite{wang2017knowledge}. An efficient implementation of these models in DGL is presented in~\cite{zheng2020dgl}.
Nevertheless, these KGE models do not naturally generalize in the few-shot scenario, where only a few edges are available for a rare edge type, which challenges learning the relation embedding. This was addressed in~\cite{chen2019meta}, where a meta-learning model is proposed to learn the relation embeddings in an inductive fashion. However, this inductive-relation KGE model require a specialized training scheme, can not learn inductive node embeddings, and can not incorporate node features if available.

Graph convolutional networks learn embeddings for nodes and edges in the graph by applying a sequence of nonlinear operations parametrized by the graph adjacency matrix and utilize node and edge features~\cite{kipf2016semi,schlichtkrull2018modeling}. An inductive implementation of these models allows for learning node embeddings in an inductive fashion~\cite{hamilton2017inductive}. The RGCN model~\cite{schlichtkrull2018modeling} has been successful in link prediction, where the RGCN is supervised by KGE models~\cite{wang2017knowledge}. However, these GCN models for link prediction inherit the limitation of the KGE models, and are challenged in learning relation embeddings for rare edges types. 

\subsection{Contributions}
 This paper addresses the aforementioned limitation of GCN models by introducing a novel inductive-RGCN that learns the relation and the node embeddings in an inductive fashion.  The proposed I-RGCN naturally addresses the few-shot link prediction and outperforms competing state-of-the-art models.  I-RGCN is also tested in the DRKG for Covid-19 drug-repurposing. The drug discovery task is naturally formulated in a few-shot learning setting.   The preliminary results indicate that several drugs used in clinical trials are discovered as possible drug candidates.  While this study, by no means recommends specific drugs, it demonstrates a powerful deep learning methodology to prioritize existing drugs for further investigation, which holds the potential of accelerating therapeutic development for COVID-19.

\section{Few-shot link prediction formulation}

Consider the heterogeneous graph with $T$ node types and $R$ relation types defined as $\mathcal{G}:=\{\{\mathcal{V}_{t}\}_{t=1}^{T},\{\mathcal{E}_r\}_{r=1}^{R}\}$. The $t$th node type is defined as $\mathcal{V}_{t}:=\{v^{t}_n\}_{n=1}^{N_t}$ and may represent  Genes or Chemical compounds in the DRKG.
The $r$th relation type holds all interactions of a certain type among $\mathcal{N}^{t}$ and $\mathcal{N}^{t'}$  $\mathcal{E}_{r}:=\{(v^{t}_n,v^{t'}_n)\in\mathcal{N}^{t}\times \mathcal{N}^{t'} \}$ and may represent that a chemical compound inhibits a gene or that a disease is treated by a chemical compound.


Consider also that each node $n_t$ is associated with a $F\times1$ feature vector $\mathbf{x}_{n_t}$. This feature may represent an embedding of the protein sequence associated with a gene~\cite{wang2017accurate}. {In KGs some node types may not have features for these we use an embedding layer to represent their features.} 

\textbf{Few shot link prediction.} Given $R-1$ sets of edges $\{\mathcal{E}_{r}\}_{r=1}^{R-1}$, a nodal attribute vector $\mathbf{x}_{n_t}$ per node $n_t$, and a small set of links in the few-shot relation $\mathcal{E}_{R}$ with $|\mathcal{E}_{R}|\le K$, the few-shot link prediction amounts to inferring the missing links of the rare type $R$. In the DRKG, this few-shot relation is for example coronavirus treatment.

\section{Learning inductive embedding for GNNs}
The relational GCN (RGCN)~\cite{schlichtkrull2018modeling} extends the graph convolution operation~\cite{kipf2016semi} to heterogenous graphs.
An RGCN model is comprised by a sequence of RGCN layers. The $l$th layer computes the $n$th node representation $\mathbf{h}^{(l+1)}_{n}$ as follows
\begin{align}
\mathbf{h}^{(l+1)}_{n}
	:=
	\sigma\left(\sum_{r=1}^{R}\sum_{ n'\in\mathcal{N}_n^{r}} 
	{\mathbf{h}}_{n'}^{(l)} \mathbf{W}^{(l)}_{r}\right)
	\label{eq:sem}
\end{align}
where $\mathcal{N}_n^{r}$ is the neighborhood of node $n $ under relation $r$, $\sigma$ the rectified linear unit non linear function, and $\mathbf{W}^{(l)}_{r}$ is a learnable matrix associated with the $r$th relation. 
Essentially, the output of the RGCN layer for node $n$ is a nonlinear combination of the hidden representations of neighboring nodes weighted based on the relation type. The node features are the input of the first layer in the model i.e. $\mathbf{h}^{(0)}_{n}=\mathbf{x}_n$, where $\mathbf{x}_n$ is the node feature for node $n$. For node types without features we use an embedding layer that takes as input an one-hot encoding of the node id. 

The RGCN model in this paper is supervised by a DistMult model~\cite{yang2014embedding} for link prediction. The loss function
\begin{align}
\label{eq:rgcnsup}
\min \sum_{n_t,{r},n_{t'} \in \mathbb{D}^+ \cup \mathbb{D}^-} \log(1+\exp(-y \times
\mathbf{h}_{n_t}^\top \mbox{diag}(\mathbf{h}_r)\mathbf{h}_{n_{t'}}))
\end{align}
where $\boldsymbol{h}^{\top}$ denotes the transpose of a matrix, $\text{diag}{(\mathbf{r})}$ denotes a diagonal matrix with $\mathbf{r}$ on its diagonal,   $\mathbf{h}_{n_t}$, $\mathbf{h}_r$, $\mathbf{h}_{n_{t'}}$ are the embedding of
the head entity $n_t$, relation ${r}$ and the tail entity $n_{t'}$, respectively and $\mathbb{D}^+$ and $\mathbb{D}^-$ are the positive and negative sets of
triplets and $y=1$ if the triplet corresponds to a positive example and $-1$ otherwise.
The scalar represented by $\mathbf{h}^{\top} \text{diag}{(\mathbf{h}_r)} \mathbf{t}$ denotes the score of triplet $({h}, {r}, {t})$ as given by the DistMult model~\cite{yang2014embedding}. The entity embeddings are obtained by the final layer of the RGCN.  The relation type embedding are trained directly from~\eqref{eq:rgcnsup}.

Such a model~\eqref{eq:rgcnsup} is vulnerable when only few training edges are available for a certain relation type. The small number of edges will challenge the learning of the embedding vector $\mathbf{r}$ for the rare relation. 

\subsection{Inductive RGCN}
Certain relation-types may be rare in the training set of links and require a  specialized architecture. To address such a few-shot scenario, we introduce a MLP to learn the relation embeddings. Consider the node embeddings $\{\mathbf{h}_{n_t}\}$ and $\{\mathbf{h}_{n_{t'}}\}$  extracted from the ultimate layer of the RGCN model where $n_{t} \in \mathcal{V}_t$ and $n_{t'} \in \mathcal{V}_{t'}$.  The proposed MLP learns an embedding for the $r$th relation as follows
\begin{align}
    \mathbf{h}_r:=\frac{1}{|\mathcal{E}_r|}\sum_{(n_t,n_{t'})\in\mathcal{E}_r}\sigma(\mathbf{W}_1\sigma(\mathbf{W}_2 (\mathbf{h}_{n_t}|\!|\mathbf{h}_{n_{t'}}))
\end{align}
where $|\!|$ denotes the vector concatenation. Note that the relation embedding $\mathbf{h}_r$ is calculated as a nonlinear function of the node embedding for all node pairs $(n_t,n_{t'})$ participating to a certain relation type $(n_t,n_{t'})\in\mathcal{E}_r$. This allows the I-RGCN to learn relation embeddings in an inductive fashion. This model is supervised by the following loss
\begin{align}
    \mathcal{L}_{\textsc{FSLP}}:= \log(1+\exp(-y \times \mathbf{h}_{n_t}^\top \mbox{diag}(\mathbf{h}_r)\mathbf{h}_{n_{t'}}))
    \label{eq:indloss}
\end{align}
where $\mathbf{h}_{n_t}^\top \mbox{diag}(\mathbf{h}_r)\mathbf{h}_{n_{t'}}$ denotes the triple score and $y$ is $-1$ for the negative triples and $1$ stands for the positive ones. We create negative triples by fixing the head node of a positive triple and randomly selecting a tail node of the same type as the original tail node.  Differently from~\eqref{eq:rgcnsup}, the relation embedding $\mathbf{h}_r$ is learned in an inductive fashion from the participating node pairs ($n_t$, $n_{t'}$). Hence, upon learning the MLP parameters $\mathbf{W}_1, \mathbf{W}_2$ the relation embedding will be computed with a forward pass. This obviates the few-shot learning hurdle and enables the model to  generalize to rare or even unseen relations. 

\section{Experiments}
\subsection{Few shot link prediction}
\begin{table}[t]
  \caption{Statistics of datasets.}
  \label{tab:dataset}
  \centering
  \scalebox{.75}{ 
  \begin{tabular}{cccc}
    \toprule
     & Nodes & Edges & Relation types \\
    \midrule
    IMDb & \specialcell{ movie : 4,278 \\  director : 2,081 \\  actor : 5,257} & 17,106& 12 \\
    \midrule
    DBLP & \specialcell{ author : 4,057 \\  paper : 14,328 \\  term : 7,723 \\  venue : 20} & 119,783& 12 \\
    \bottomrule
  \end{tabular}}
\end{table}
\textbf{Baselines.}We consider the state-of-the-art KGE models RotatE~\cite{sun2019rotate}, ComplEx~\cite{trouillon2016complex}, and the RGCN model~\cite{schlichtkrull2018modeling} as baselines for comparison. The parameters of these methods have been optimized via cross validation.
\begin{table*}[ht!]
\caption{Experiment results (\%) on the IMDb dataset for k-shot link prediction. }
\label{tab:link_pred_dif_splits}
\centering
\scalebox{.75}{  
\begin{tabular}{|c|c|c|c|c|c|c|c|c|c|c|c|c|}
\hline
   &\multicolumn{4}{c|}{MRR}  &
 \multicolumn{4}{c|}{ Hit 1}      &\multicolumn{4}{c|}{ Hit 10}\\ \hline
K &ComplEx&RotatE& RGCN  & I-RGCN  &ComplEx&RotatE& RGCN & I-RGCN  &ComplEx&RotatE& RGCN &I-RGCN         \\ \hline
  10       &6.88 &11.80 & 1.32 &\bf33.56    &1.75&6.74 &0.13 &\bf25.32     &14.25&18.35 & 1.07 & \bf53.55 \\ \hline
 50       &8.48&12.56 &15.26&\bf53.24  &3.34&7.76 &7.42 & \bf45.14 &15.70&19.12  &28.88 &\bf69.32\\\hline
 100       &8.61&12.57 &18.78 & \bf53.63    &3.44&7.86&9.59 &\bf40.27     &15.71&18.40 &36.84&\bf77.38  \\\hline
1000       &68.37& 70.09&95.23 & \bf96.06    &65.48&67.52&91.72 &\bf93.56 &72.23&73.50 &99.72 &\bf99.81 \\ \hline
\end{tabular}}
\end{table*}
We use the IMDB and DBLP datasets~\cite{fu2020magnn} detailed in Table~\ref{tab:dataset}. The total number of edges in the few-shot relation are $1559$ for the IMDB and $3534$ for the DBLP.  In the experiments. we train with only  $K$ links from the few-shot relation and all the links from the other relations and test on the rest edges of the few-shot relation, which are $|\mathcal{E}_{R}|-K$. 
The nodes in the IMDB and DBLP graphs are associated with feature vectors. Further, information on the datasets is included in the Appendix. 

Tables~\ref{tab:link_pred_dif_splits} and~\ref{tab:link_pred_dif_splits_dblp} report the MRR, Hit-1 and Hit-10 scores of the baseline methods along with the inductive RGCN and the RGCN in the task of few-shot link prediction for the IMDB and DBLP datasets respectively. The I-RGCN significantly outperforms the alternative methods in the task of few-shot link prediction. Specifically, for $K$=10 the MRR of the inductive method is one order of magnitude greater. This corroborates the advantage of the inductive relation learning for the few-shot learning task. As the number of training edges increases at $K$=1000, it is observed that the RGCN performance approaches the performance of the I-RGCN. This suggests that the I-RGCN method performs well also in non few-shot learning tasks. The worse performance of KGE models is explained since these do not account for node features and do learn inductive relation embeddings.

To further validate the performance of the I-RGCN we conduct a general link prediction evaluation by splitting the links in training, validation, and testing at random \emph{irrespective} of their relation type. The results for different percentages of training links are reported in Table~\ref{tab:link_pred_dif_splits_gen}. I-RGCN outperforms even in this training scenario the RGCN and KGEs baselines, which further corroborates the efficiency of the model.

\begin{table*}[ht!]
\caption{Experiment results (\%) on the DBLP dataset for k-shot link prediction. }
\label{tab:link_pred_dif_splits_dblp}
\centering
\scalebox{.75}{  
\begin{tabular}{|c|c|c|c|c|c|c|c|c|c|c|c|c|}
\hline
   &\multicolumn{4}{c|}{MRR}  &
 \multicolumn{4}{c|}{ Hit 1}      &\multicolumn{4}{c|}{ Hit 10}\\ \hline
K &ComplEx&RotatE& RGCN  & I-RGCN  &ComplEx&RotatE& RGCN & I-RGCN   &ComplEx&RotatE& RGCN &I-RGCN         \\ \hline
  10       &5.97&7.41 & 7.42 &\bf27.95    &1.37&2.13 &2.95 &\bf13.78     &11.81&15.36 & 12.51 & \bf60.51 \\ \hline
 50       &6.32&7.87&17.25&\bf83.42  &1.55&2.26&10.95 & \bf72.54 &12.59&16.64  &26.84 &\bf96.82\\\hline
 100       &7.24&10.66 &32.45 & \bf90.00    &1.99&04.43&23.46 &\bf85.27     &14.96&20.92 &49.85&\bf97.61  \\\hline
1000       &36.56& 46.51&91.34 & \bf96.82    &30.62&39.83&86.43 &\bf94.41 &46.45&59.27 &98.59 &\bf99.81 \\ \hline
\end{tabular}}
\end{table*}
\begin{table*}[ht!]
\caption{Experiment results (\%) on the IMDb dataset for link prediction. }
\label{tab:link_pred_dif_splits_gen}
\centering
\scalebox{.75}{  
\begin{tabular}{|c|c|c|c|c|c|c|c|c|HHHHc|c|c|c|}
\hline
 Metrics  &\multicolumn{4}{c|}{MRR}  &
 \multicolumn{4}{c|}{ Hit 1}  &\multicolumn{4}{H}{ Hit 3}    &\multicolumn{4}{c|}{ Hit 10}\\ \hline
Training links &ComplEx&RotatE& RGCN  & I-RGCN  &ComplEx&RotatE& RGCN &  I-RGCN  &ComplEx&RotatE& RGCN &  I-RGCN   &ComplEx&RotatE& RGCN &  I-RGCN         \\ \hline
  95\%       &94.15 &93.97 & 93.38 &\bf95.12    &\bf93.75&93.48 &89.31 &92.75       &94.12&94.04 &97.25 & \bf98.59 &94.74&94.56 & \bf99.29 & 98.24 \\ \hline
 90\%       &88.87&88.89 &89.30 &\bf93.98   &88.25&87.87 &83.45 & \bf91.52       &88.67& 89.07&94.68&\bf96.34&89.74&90.35  &97.66 &\bf98.12\\\hline
 80\%       &78.55&78.90 &83.46 & \bf90.03    &77.45&76.83 &76.59 &\bf86.70        &78.16& 79.28 &88.58&\bf 93.76 &79.96&81.94 &94.72&\bf95.93  \\\hline
 70\%       &69.59& 69.56&82.73 & \bf87.00    &67.98&66.73 &76.09 &\bf 82.95        &68.92& 70.05&87.61 &\bf90.76&71.89&73.49 &93.76 &\bf94.07 \\ \hline
 60\%       &60.40&60.90 &78.16 & \bf81.53    &58.49&57.60 &70.63 &\bf77.01        &59.54& 61.17&83.36&\bf86.87&62.92&65.57 &\bf91.14&90.27 \\ 
 \hline
\end{tabular}}
\end{table*}
\subsection{Drug-repurposing via I-RGCN}
For this experiment we will utilize the drug-repurposing knowledge graph (DRKG) constructed in~\cite{drkg2020}. The DRKG collects interactions from a collection of biological databases such as Drugbank~\cite{drugbank@2017}, GNBR~\cite{percha2018global},  Hetionet~\cite{hetionet@2017}, STRING~\cite{string@2019}, IntAct~\cite{intact} and DGIdb~\cite{dgidb@2017}.

Drug-repurposing aims at discovering the most effective existing drugs to treat a certain disease. Drug-repurposing can be formulated as predicting direct links in the DRKG such as predicting whether a drug treats a disease or as predicting whether a compound inhibits a certain gene which is related to the target disease.   Drug-repurposing  can be viewed as a few-shot link prediction task since only a few edges are available related to novel diseases in the DRKG. %

We use corona-virus related diseases, including SARS, MERS and SARS-COV2, as target diseases representing Covid-19 as their functionality is similar. 
We aim at predicting links among gene entities associated with the target disease and drug entities.
We select FDA-approved drugs in Drugbank as candidates, while we exclude for simplicity drugs with molecule weight less than 250 daltons, as many of certain drugs are actually health drugs. 
This amounts to 8104 candidate drugs.

 We also obtain 442 Covid-19 related genes from the relations extracted from~\cite{gordon2020sars,zhou2020network}. Similarly, we obtain the node embeddings for the gene and drugs, and the embeddings for the corresponding relations. Next, we score all triples and rank them per target gene. This way we obtain 442 ranked lists of drugs. Finally, to assess whether our prediction is in par with the drugs used for treatment, we check the overlap among the top 100 predicted drugs and the drugs used in clinical trials per gene. We used 32 clinical trial drugs for Covid-19 to validate our predictions\footnote{The clinical trial drugs were collected from \url{http://www.covid19-trails.com/}}. Table~\ref{tab:predicted_drugs_gene} lists the clinical drugs included in the top-100 predicted drugs across all the genes with their corresponding number of hits for the RGCN and I-RGCN.  It can be observed, that several of the widely used drugs in clinical trials appear high on the predicted list, and that I-RGCN shows a higher hit rate than RGCN. Hence, the inductive relation prediction module is more appriopriate in predicting links when information about the nodes is limited, such as is the case with the novel Covid-19 disease node.

\begin{table}[]
    \centering
       \caption{Drug inhibits gene scores for Covid-19. Note that a random classifier will result to approximately 5.3 per drug. This suggests that the reported predictions are significantly better than random.}
    \label{tab:predicted_drugs_gene}
        \scalebox{0.8}{  
    \begin{tabular}{cc|cc}\toprule
    \multicolumn{2}{c|}{I-RGCN}& \multicolumn{2}{c}{RGCN}\\
         Drug name& \# hits &Drug name& \# hits\\\midrule
	Dexamethasone&	240&Chloroquine &	69\\
	Ribavirin&	142&Colchicine&	41\\
	Colchicine&	128	&Tetrandrine&	40\\
	Chloroquine	&115&Oseltamivir&	37\\
	Methylprednisolone&	86&Azithromycin&	36\\
	Tofacitinib	&75&	Tofacitinib	&33\\
	Thalidomide&	70&		Ribavirin&	32\\
	Losartan&	64&Methylprednisolone&	30\\
	Hydroxychloroquine&	48&	Deferoxamine&	30\\
	Oseltamivir	&46&	Thalidomide&	25\\
	Deferoxamine&	34&Dexamethasone&	24	\\
	Ruxolitinib&	23&Bevacizumab&	21\\
	Azithromycin&	23&	Hydroxychloroquine&	19\\
	Nivolumab&	11&	Losartan&	19\\
	Tradipitant&	11&		Ruxolitinib&	13\\
	Bevacizumab	&10&Eculizumab&12\\
	Eculizumab&	7&Tocilizumab&	11\\
	Baricitinib&	6&	Anakinra&	11\\
	Sarilumab&	6&Sarilumab	&8\\
	Tetrandrine&	6&Nivolumab&	6\\
         \bottomrule
\end{tabular}}
\end{table}

\section{Conclusion}
In this paper we develop a novel I-RGCN that learns inductive relation embeddings and can be applied for few-shot link prediction and drug repurposing. I-RGCN consistently outperforms baseline models in the IMDB and DBLP datasets for few-shot link prediciton. We also formulate the Covid-19 drug-repurposing task as a link prediction over the DRKG. I-RGCN successfully identifies a subset of clinical trial drugs for Covid-19 and can be used to assist researchers and  prioritize existing drugs for further investigation in the Covid-19 treatment.

\bibliography{my_bibliography.bib}
\bibliographystyle{icml2020}
\appendix
\section{Datasets}
We use the IMDB and DBLP datasets~\cite{fu2020magnn} detailed in Table~\ref{tab:dataset}, where the third column denotes the total size of edges in the few-shot relation that is $|\mathcal{E}_{R}|$. The nodes in the IMDB and DBLP graphs are associated with feature vectors. The original datasets in~\cite{fu2020magnn} are used for node classification. We adapt the datasets and create new edge types, where the edges are parametrized by the label of the associated nodes. For example, the edge type (director, directed, movie) becomes (director, directed\_drama, movie) if the associated movie is in the drama genre, and the same transformation undergoes the (actor, played, movie) relation. Since, there are 3 labels for movies, this way the original 4 edge types become 12. The same transformation happens in the DBLP dataset. 

\end{document}